\documentclass[letterpaper, 10 pt, conference]{ieeeconf}  

\IEEEoverridecommandlockouts                              

\usepackage{graphics} 
\usepackage{epsfig} 
\usepackage{amsmath} 
\usepackage{multirow}
\usepackage{cite}

\title{\LARGE \bf
Impedance Control of a Transfemoral Prosthesis using Continuously Varying Ankle Impedances and Multiple Equilibria
}

\author{Namita Anil Kumar$^{1}$, Woolim Hong$^{1}$, and Pilwon Hur$^{2}$
\thanks{*This work was not supported by any organization}
\thanks{$^{1}$Namita Anil Kumar and Woolim Hong are with the Department of Mechanical Engineering, Texas A\&M University, College Station, TX 77843, USA
        {\tt\small \{namita.anilkumar,ulim8819\}@tamu.edu}}%
\thanks{$^{2}$Pilwon Hur is with faculty of the Department of Mechanical Engineering,  Texas  A\&M  University,  College  Station,  TX  77843,  USA
        {\tt\small pilwonhur@tamu.edu}}%
}

\begin{document}

\maketitle
\thispagestyle{empty}
\pagestyle{empty}

\begin{abstract}

Impedance controllers are popularly used in the field of lower limb prostheses and exoskeleton development. Such controllers assume the joint to be a spring-damper system described by a discrete set of equilibria and impedance parameters. Said parameters are estimated via a least squares optimization that minimizes the difference between the controller's output torque and human joint torque. Other researchers have used perturbation studies to determine empirical values for ankle impedance. The resulting values vary greatly from the prior least squares estimates. While perturbation studies are more credible, they require immense investment. This paper extended the least squares approach to reproduce the results of perturbation studies. The resulting impedance parameters were successfully tested on a powered transfemoral prosthesis, AMPRO II. Further, the paper investigated the effect of multiple equilibria on the least square estimation and the performance of the impedance controller. Finally, the paper uses the the proposed least squares optimization method to estimate knee impedance.

\end{abstract}

\section{INTRODUCTION}
\begin{figure}[b]
      \centering
      \framebox{\centering\includegraphics[width = 3.3in]{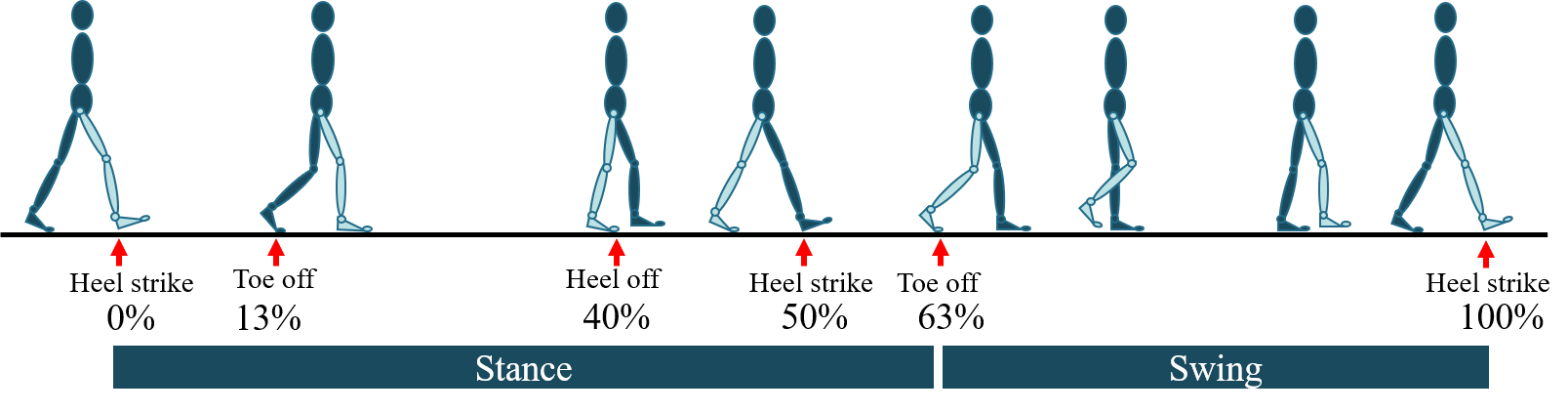}}
      \caption{Gait cycle with important kinematic changes}
      \label{fig:gaitcycle}
\end{figure}

The field of prosthesis design has been growing considerably over the past years, addressing the needs of both transtibial and transfemoral amputees \cite{Perry1997,Blumentritt1997,Romo2000}. Upon understanding the limitations of passive prostheses, researchers have made strides to develop powered prostheses \cite{Au2007,Lenzi2017,Martinez-Villalpando2008,Sup2008,Thatte2014,Windrich2016,Zhao2017a,Azimi2017,Elery2018,Azocar2018}. Said prostheses implement control strategies that fall into two major groups: impedance controllers that attempt to mimic human joint impedance \cite{Sup2008,Fey2014} and trajectory-tracking controllers that follow optimized joint trajectories \cite{Gregg2014,Paredes2016,Martin2017,Zhao2017a}. Of the two classes, the former has displayed greater promise in mimicking human-like gait kinetics. As stated in \cite{Sup2008}, an impedance controller enables the user to interact with the device much like in the case of healthy walking. An impedance controller consists of parameters pertaining to stiffness, damping and the equilibrium angle of the joints. By modulating these parameters, the joint torque required for support and propulsion of the human body can be generated. According to \cite{Sup2008}, researchers sectioned the gait cycle into 4-6 phases based on kinematic changes observed in a healthy human gait cycle (refer Fig. \ref{fig:gaitcycle}). 
Each phase has a set of three constant values--stiffness, damping, and the equilibrium angle. These values were initially estimated using a least squares optimizer that reduced the error between the torque of the impedance controller and human torque data \cite{Winter2009}. During testing, these estimates were tuned. Though successful, this approach mandated the manual tuning of several parameters. The study \cite{Eilenberg2010} implemented a series elastic actuator to modulate the impedance of the ankle joint in a transtibial prosthesis. Assuming the ankle to be a spring-damper system, the study estimated the stiffness parameters using a least squares optimization approach. The study, \cite{Blaya2004}, proposed an ankle-foot exoskeleton that aided stroke patients in combating foot-drops. The system was manipulated using an impedance controller that was automatically tuned using a simple algorithm. Other notable attempts at estimating joint impedance during walking are \cite{Shamaei2013b} and \cite{Hu2016}. The former proposes  quasi-stiffness, which estimates joint stiffness by calculating the slope of the torque vs. angle curve during stance. The latter solved a constrained optimization problem where the joint impedance was the decision variable and the dynamics of a humanoid bipedal served as constraints. A common attribute of these studies is that the estimated impedance does not vary smoothly and continuously throughout the gait cycle. This paper will refer to the above group of estimation methods as \textit{theoretical approaches}.

In the field of science, the most accepted method of identifying a system's parameters is by experimentally inducing perturbations. This paper will refer to such a methods as \textit{empirical approaches}. With the objective of empirically determining the ankle impedance while walking, researchers conducted experimental studies on the ankle \cite{Rouse2014,Lee2016a,Shorter2018}. These studies perturbed the ankle at various instances of the gait cycle. The ankle's response to the perturbation was gathered and analyzed to evaluate empirical values for stiffness and damping. It was reported that the ankle stiffness increases upon heel-strike until terminal-stance, following which the stiffness reduces until toe-off and maintains an almost constant value during swing phase. The damping parameter was observed to be high during heel-strike and toe-off. Unlike the results of the prior \textit{theoretical approaches}, the \textit{empirical} impedance parameters varied continuously and smoothly throughout the gait cycle. The first objective of this paper is to bridge the gap between the \textit{theoretical} and \textit{empirical approaches} by extending the least squares method proposed by \cite{Sup2008} to produce results that conform to \cite{Rouse2014,Lee2016a,Shorter2018}. Specifically, the stiffness and damping parameters will be allowed to vary continuously throughout the gait cycle.

In \cite{Fey2014}, researchers successfully implemented the \textit{empirical} impedance parameters reported in \cite{Rouse2014} to control the ankle of a transfemoral prosthesis. Though the study enforced impedance control on the knee, it utilized quasi-stiffness as the impedance parameter. Unfortunately, to the authors' knowledge, there have been no published attempts at \textit{empirically} estimating knee or hip impedance during the gait cycle via perturbation studies. Any insight gained in this matter is limited to the swing phase \cite{Koopman2016}. Possible reasons for this gap in knowledge are (i) the huge investment required to conduct perturbation studies, and (ii) the grand challenge of isolating of the effects induced by the perturbation to the joint being studied. Thus, the research community would highly benefit from \textit{theoretical approaches} to estimating impedance. It is hoped that the least squares optimization method proposed in this study can be used to estimate knee and hip impedance. The concluding section of this paper presents a preliminary estimate of knee impedance. The resulting impedance will be compared with other literary works.

A recent study by \cite{Mohammadi2019} proposed a continuum of equilibria in contrast to the discrete set of equilibria implemented in \cite{Sup2008}.  The study also estimated the impedance of the knee joint using a \textit{theoretical approach}. This study has raised questions regarding the effect of multiple equilibria on the performance of impedance controllers. The second objective of this paper is to fill this gap in knowledge by investigating the role of multiple equilibria on the proposed least squares estimation method. The validity of the resulting impedance will be assessed via implementation on an existing prosthesis AMPRO II. Additionally, attempts will be made to reduce the required tuning process during implementation.

\section{\uppercase{Least squares estimation of impedance parameters}}
The optimization problem solved in this paper is fundamentally similar to the one used by \cite{Sup2008}. The lower limb joints are modeled as a spring-damper system. Let $K$ and $D$ represent the stiffness and damping of the joint, respectively. The generated torque can be calculated as follows. 
\begin{equation}
    \tau = K (\theta - \theta_{eq}) + D \dot{\theta}
    \label{Eq:ImpedControl}
\end{equation}
In (\ref{Eq:ImpedControl}), $\theta$ and $\dot{\theta}$ signify the position and velocity of the joint, while $\theta_{eq}$ is the equilibrium angle of the joint. It is desired that the generated torque be similar to that found in healthy human walking \cite{Winter2009}, say $\tau_{data}$. Thus, the optimization problem minimizes the error between the torque $\tau$ and $\tau_{data}$. Per \cite{Rouse2014,Lee2016a,Shorter2018}, the stiffness and damping parameters continuously vary throughout the gait cycle in a smooth manner. Most of the variation in the impedance parameters are observed during the stance phase, while the parameters adopt an almost constant value during the swing phase. To permit the continuous variation of stiffness and damping, while maintaining minimal decision variables, the stiffness and damping parameters were represented by polynomials during the stance phase. The orders of the polynomials were adjusted to get a better fit (i.e. reduce difference between $\tau$ and $\tau_{data}$). During the swing phase the impedance parameters were assigned constant values: $k_{swing}$ and $d_{swing}$. Supposing $m$ and $n$ represent the order of the stiffness and damping polynomials, the impedance parameters at any instant during the gait cycle are determined as follows.
\begin{align}
K(t)=\begin{cases}
	\sum_{i=0}^{m}k_i t^i &\text{for} ~ 0 \leq t < 0.63 \\
	k_{swing} &\text{for} ~ 0.63 \leq t \leq 1
\end{cases} \\
D(t) = \begin{cases}
	\sum_{i=0}^{n}d_i t^i &\text{for} ~ 0 \leq t < 0.63 \\
	d_{swing} &\text{for} ~ 0.63 \leq t \leq 1
\end{cases}
\end{align}
Per the requirement for continuity in the impedance parameters, $k_{swing} = k_0$ and $d_{swing}=d_0$. Much like \cite{Sup2008}, the gait is sectioned based on kinematic changes; making $\theta_{eq}$ a set of angles. The optimization problem can be summarized as follows.
\begin{align}
	\underset{\theta_{eq},k_i,d_i}{\min} \quad & \Vert \tau_{data}-\tau \Vert_2 \label{Eq:cost}\\
	\text{Subject to:} \quad&	K(t) \geq 0 \qquad \qquad D(t)  \geq 0 \label{Eq:Posit}\\
	& K(0) = K(1) \quad\quad  D(0) = D(1) \label{Eq:Continity}\\
	& |\Delta\tau/\Delta t| \leq c \label{Eq:Lip}
\end{align}
The constraints listed in (\ref{Eq:Posit}) force the positivity of the impedance parameters. Further, (\ref{Eq:Continity}) ensures that the parameters maintain continuity between gait cycles. The last constraint, (\ref{Eq:Lip}), forces the resulting $\tau$ to be continuous using a Lipschitz constant, $c$. Additional bounds were added, as needed, to restrict the value of the equilibrium angles.  

\subsection{Multiple Equilibria}
\begin{table}[t]
\caption{The four sets of multiple equilibria that resulted from sectioning the gait are as follows.}
\label{tab:eq_sets}
\begin{center}
\begin{tabular}{c|c|c|c|c}
\hline
\hline
\textbf{Set label} & \multicolumn{4}{c}{\textbf{Sections of the gait cycle}}\\
\hline
\hline
Set A   & 0\% - 13\%  & 13\% - 40\% & 40\% - 63\%   & 63\% - 100\%\\ 
\hline
Set B   & \multicolumn{2}{|c|}{0\% - 40\%} & 40\% - 63\%   & 63\% - 100\%\\ 
\hline
Set C   & \multicolumn{3}{|c|}{0\% - 60\%} & 63\% - 100\%\\ 
\hline
Set D   & \multicolumn{4}{|c}{0\% - 100\%}\\
\hline
\hline
\end{tabular}
\end{center}
\end{table}

To study the impact of multiple equilibria on the impedance controller, four sets of equilibria were established. The first set echoes the one found in \cite{Sup2008}. The gait is sectioned in accordance to the foot contact sequence during stance phase. The first phase initiates at heel-strike (0\%) and continues until foot-drop (13\%), followed by the second phase that terminates at heel-off (40\%). The third phase exists between heel-off and toe-off (63\%). Unlike \cite{Sup2008}, the swing phase of the gait cycle is not sectioned. This set of equilibria has been labeled as \textit{Set A}. One of the objectives of this paper is to reduce the number of impedance control parameters that require tuning. The number of equilibria contributes heavily towards the number of tuning parameters. Thus, this paper wishes to determine the minimum number of equilibria required to generate natural human-like walking (both kinematically and kinetically). The remaining three sets of equilibria implement fewer sections of the gait cycle. Table \ref{tab:eq_sets} lists said sets. The sectioning proposed in \textit{Set B} is similar to the one proposed by \cite{Rouse2014}.   

\subsection{Results of the optimization}
\begin{figure}[tb]
      \centering
      \framebox{\parbox{3.3in}{\centering\includegraphics[width=3.25in]{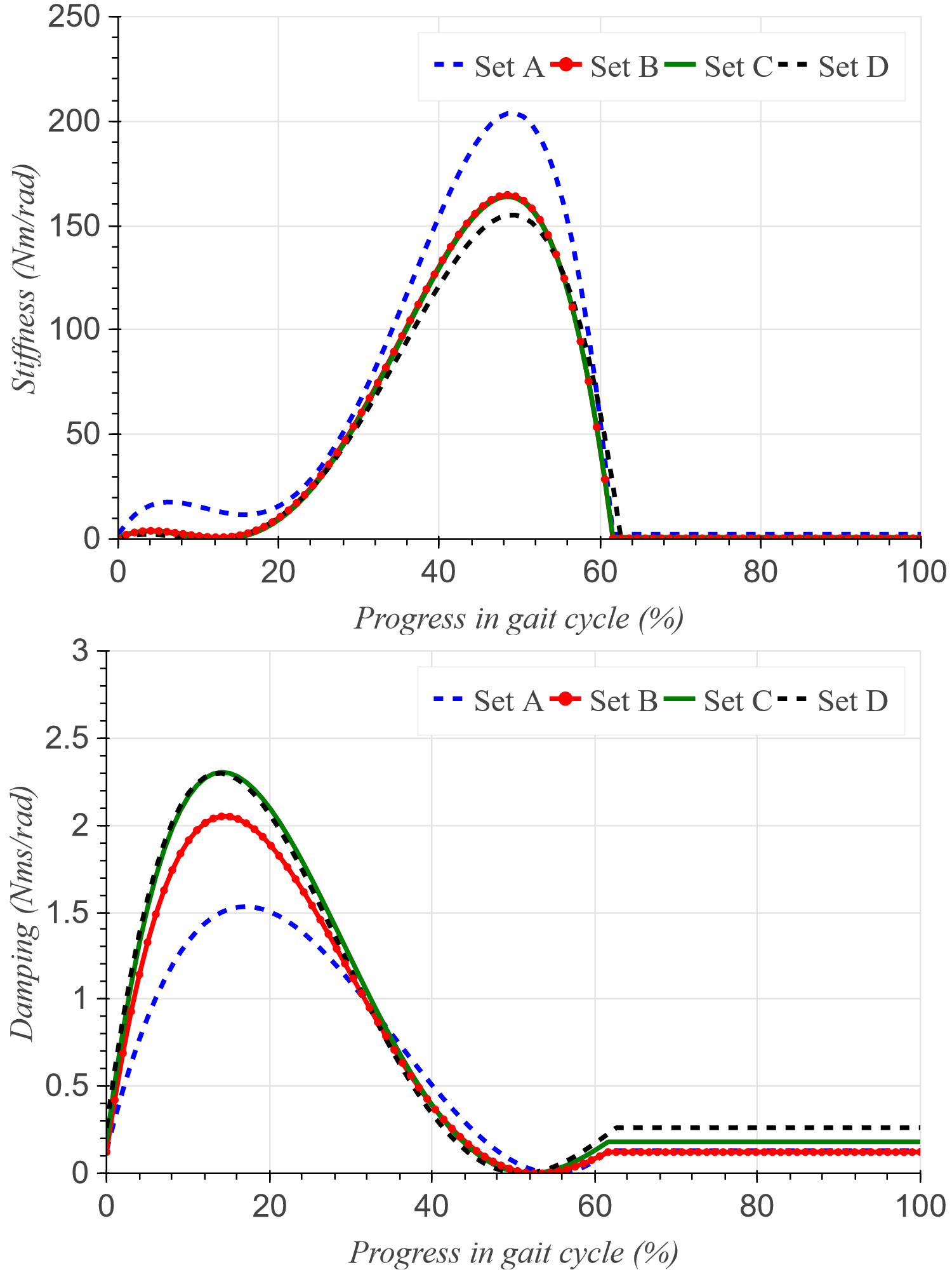}
        }}
      \caption{Optimization results. \textbf{Top}: Stiffness curves, \textbf{Bottom}: Damping curves. The associated polynomial coefficients have been included in the Appendix (Table \ref{tab:polyCoeff}).}
      \label{fig:OptimResults}
\end{figure}
The minimum order of the stiffness and damping polynomials required to lower the optimal cost was determined to be 4. This study fixed the order of the stiffness and damping polynomial to be the same. It was observed that the trend of the stiffness and damping parameters was not sensitive to the equilibria set. Fig. \ref{fig:OptimResults} depicts the stiffness and damping parameters. The corresponding torques, $\tau$, has been presented in the Appendix. It is evident that all sets of parameters attained a suitable cost to the optimization problem. While the trend of the stiffness parameter, during stance, resembled that reported by \cite{Rouse2014,Lee2016a,Shorter2018}, the values are greatly lower. The trend of the damping parameters, on the other hand, did not entirely conform to the results presented in \cite{Rouse2014,Lee2016a,Shorter2018}. Though it portrayed high damping post heel-strike, there is little to no damping during terminal-stance. Better results could be attained by increasing the order of the damping parameter. Table \ref{tab:optim_eq_sets} presents the equilibrium angles that resulted from the optimization. \textit{Sets A} to \textit{C} showed similarities by having the ankle plantar-flexed during terminal-stance and dorsi-flexed during swing. The equilibrium angle for \textit{Set D} resembled a foot-drop condition--the state a human foot would conform to when physically unconstrained. It was anticipated that the foot-drop condition would pose a challenge during swing phase. It is likely that certain compensatory actions will be needed to assure sufficient foot clearance during swing.

\begin{table}[t]
\caption{Sets of multiple equilibria in radians resulting from the optimization}
\label{tab:optim_eq_sets}
\begin{center}
\begin{tabular}{c|c|c|c|c}
\hline
\hline
\textbf{Set label}   & 0\% - 13\%  & 13\% - 40\% & 40\% - 63\%   & 63\% - 100\%\\
\hline
\hline
Set A   & 0.0294  & -0.3428 & -0.3491   & 0.3029\\ 
\hline
Set B   & \multicolumn{2}{|c|}{-0.4258} & -0.4363   & 0.0000\\ 
\hline
Set C   & \multicolumn{3}{|c|}{-0.4363} & 0.1453\\ 
\hline
Set D   & \multicolumn{4}{|c}{-0.4655}\\
\hline
\hline
\end{tabular}
\end{center}
\end{table}

\section{\uppercase{Testing methodology}}

\begin{figure}[b]
      \centering
      \framebox{\centering\includegraphics[scale=0.43]{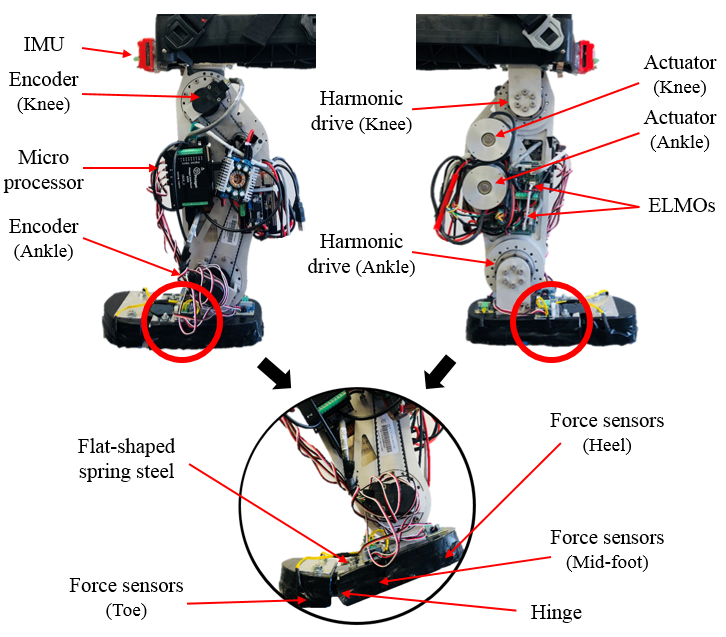}}
      \caption{AMPRO II--a custom-built powered transfemoral prosthesis}
      \label{fig:ampro2}
\end{figure}

The proposed sets of impedance parameters were tested on a custom-built powered transfemoral prosthesis shown in Fig. \ref{fig:ampro2}. AMPRO II (Fig. \ref{fig:ampro2}) has an actuated ankle and knee joint, and a passive spring-loaded toe joint. While the proposed impedance controller was implemented at the ankle, a previously published controller--a hybrid of impedance and trajectory tracking--was used to manipulate the knee. The latter has been discussed in \cite{Hong2019}. The prosthesis was operated under a time-based scheme which utilizes a parameter that linearly increases from 0 to 1 as the gait progresses from 0\% to 100\%. This parameter is used to identify the progress in the gait cycle. A force sensor placed under the heel was used to initialize the parameter. To the authors' knowledge, current state-based control schemes have limitations that are yet to be overcome \cite{Rezazadeh2018,Hong2019}. For instance, the study \cite{Rezazadeh2018} investigated the usage of thigh angle as an indicator (or phase variable) of gait progression. It was demonstrated that the resulting phase variable does not display the ideal linear behavior during mid-stance, making state identification difficult. Since the focus of this study was to evaluate the performance of continuously varying impedance parameters, it was preferred the performance be unaffected by the possible inaccuracies of state-based control. Further, it was unclear whether the stiffness at AMPRO II's toe joint would impact the ankle's performance. To study the effect of each impedance controller, in an isolated manner, the toe joint was restrained using a rigid element. Future studies will investigate the impact of toe stiffness on the generated gait.

\subsection{Experimental protocol} 
To validate the proposed idea, an indoor experiment was designed using the aforementioned powered prosthesis in Fig. \ref{fig:ampro2}. A healthy young subject (male, 5'7'' height, 150 lb weight) participated in the experiment using an L-shape simulator that helped emulate prosthetic walking. The subject was asked to walk on a treadmill at his preferred walking speed (0.7 m/s). The subject's safety was assured by handrails located on either side of the treadmill. The experiment protocol has been reviewed and approved by the Institutional Review Board (IRB) at Texas A\&M University (IRB2015-0607F).

\subsection{Tuning}\label{Tuning} 
The subject recruited showed a considerable height difference between his limbs while wearing the prosthesis. In an attempt to solve this issue, the subject was asked to wear boots during the experiments; unfortunately, the difference in height still persisted. This height difference significantly limited the amount ankle dorsi-flexion observed in the prosthetic during mid-stance. In compensation, the equilibrium angles were tuned to reduce the magnitude of the plantar-flexed angles. The tuned equilibrium angles have been documented in Table \ref{tab:tuned_eq_sets}. In addition, the stiffness and damping curves were scaled down by factors $\alpha$ and $\beta$, respectively. A major drawback of scaling was that the stiffness during swing phase was no longer sufficient to transition from the plantar-flexed equilibrium angle during terminal-stance to the swing dorsi-flexion angle. Thus, a constant stiffness term ($\gamma$) was uniformly added to the stiffness curve. The following equations describe the tuning process.
\begin{align}
    K_{tuned}(t) &= \alpha  K(t) + \gamma\\
    D_{tuned}(t) &= \beta  D(t)
    \label{Eq:TunedImped}
\end{align}
The scaling factors were reduced until certain dorsi-flexion was observed during the mid-stance phase. The term $\gamma$ was increased until the ankle displayed dorsi-flexion during the swing phase. Note that $\gamma$ was not required for \textit{Set D} since the equilibrium angle remained constant throughout the gait cycle. The stiffness curve for \textit{Set A} was scaled by a factor of $\alpha = 0.4$, while $\alpha = 0.5$ for the remaining sets. Further, while $\beta = 0.2$ for \textit{Set A}, $\beta = 0.166$ for the remaining sets. The constant term $\gamma$ was equal to 20 for all sets. Though \textit{Set D} did not necessitate a constant term, the term was implemented in the interest of conducting a systematic study where the stiffness curves of each set were approximately of same magnitude. 

\begin{table}[t]
\caption{The tuned sets of equilibria in radians}
\label{tab:tuned_eq_sets}
\begin{center}
\begin{tabular}{c|c|c|c|c}
\hline
\hline
\textbf{Set label}   & 0\% - 13\%  & 13\% - 40\% & 40\% - 63\%   & 63\% - 100\%\\
\hline
\hline
Set A   & 0.0100  & -0.0875 & -0.3490   & 0.0873\\ 
\hline
Set B   & \multicolumn{2}{|c|}{-0.1745} & -0.2617   & 0.0000\\ 
\hline
Set C   & \multicolumn{3}{|c|}{-0.2617} & 0.1452\\ 
\hline
Set D   & \multicolumn{4}{|c}{-0.2617}\\
\hline
\hline
\end{tabular}
\end{center}
\end{table}

\section{\uppercase{Results and Discussion}}
\begin{figure*}[h]
      \centering
      \framebox{\parbox{0.97\textwidth}{\centering\includegraphics[width = 0.97\textwidth]{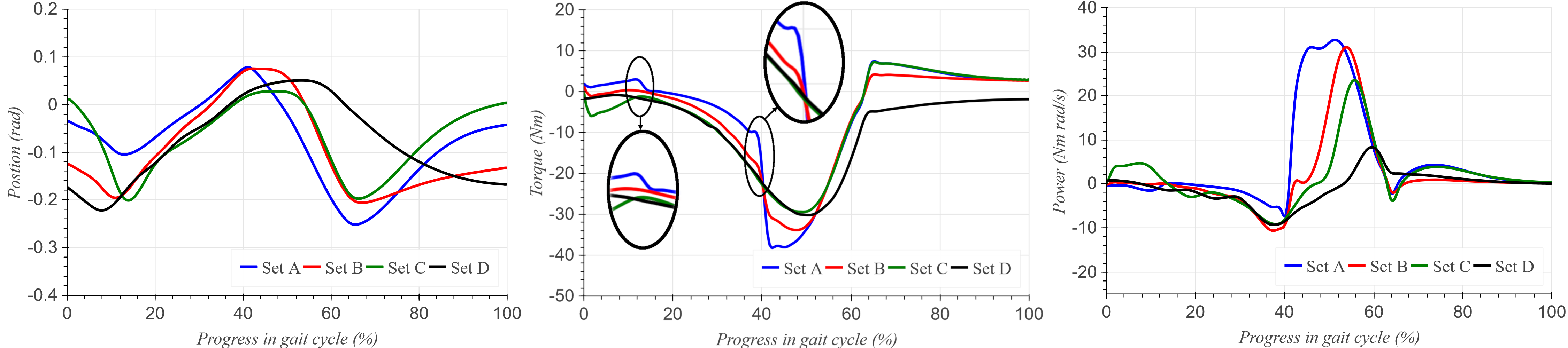}
        }}
      \caption{Averaged results of the experiments. \textbf{Left:} Ankle angle, \textbf{Middle:} Ankle torque, and \textbf{Right:} Ankle power. The sections of the the torque curve corresponding to foot-drop and heel-off have been enlarged.}
      \label{fig:ExptResults}
\end{figure*}
Figure \ref{fig:ExptResults} presents the average angular trajectory, torque, and power of the ankle for all four sets of impedance parameters. The averaged values represent 15-20 consecutive gait cycles. The standard deviation was well bounded, indicating the consistency of the observed results. Barring \textit{Set D}, the trend of the kinematic and dynamic curves are similar across the impedance sets. The trend also bears resemblance to healthy human data reported in \cite{Winter2009}. On the other hand, the magnitude of the results was greatly lower in comparison. It is strongly believed that the height difference explained in Section \ref{Tuning} is the key reason behind these discrepancies. The following subsections compare the performance of each impedance set in terms of kinematics and dynamics of the generated gait. Following which the limitations of this study have been discussed.

\subsection{Comparison of kinematics}\label{kinematic}
It should be emphasized that the stiffness and damping curves portrayed similar trends across all sets of impedance. Thus, the kinematics of the generated gait was dictated by the equilibrium angles. The following observations form the basis of this claim: (i) \textit{Set A} displayed lesser plantar-flexion proceeding heel-strike in comparison to the other sets owing to the dorsi-flexed equilibrium angle between heel-strike and foot-drop (ii) Lesser dorsi-flexion was observed during terminal-stance in \textit{Set C} and \textit{D}. Unlike these sets, \textit{Set A} and \textit{B} increase the plantar-flexed equilibrium angle in increments. It is likely that such an incremental ascension assisted the subject in achieving higher dorsi-flexion during mid and terminal-stance (iii) The variance in ankle angles, among sets, at the beginning and end of the gait cycle is due to varying swing equilibrium angles. It would be beneficial to implement a higher swing equilibrium angle since it ensures sufficient foot clearance during swing (iv) Plantar-flexion at push off was greater in \textit{Set A} due to the higher equilibrium angle. Additionally, \textit{Set A} showed an earlier descent from dorsi-flexion to plantar-flexion at heel-off (40\%). A plausible explanation is that the combined effect of heightened stiffness and higher plantar-flexion forced an earlier push-off (v) Evidently, \textit{Set D} showed an absence of dorsi-flexion during swing phase due to the plantar-flexed equilibrium angle. As anticipated, the foot-drop equilibrium angle of \textit{Set D} resulted in few stumbles during the swing phase \cite{Rosenblatt2014}.

A kinematic abnormality that cannot be overlooked is the absence of push-off in \textit{Set D}. As stated earlier, the impedance control strategy was implemented using a time parameter that linearly increased as the gait progressed. With that being said, the success of time-based control heavily depends on the subject's ability to synchronize his/her gait with said time parameter. This synchronization task proved to be a mighty challenge while testing \textit{Set D}. Specifically, the constant foot-dropped equilibrium angle introduced gait abnormalities such as exaggerated hip extension at toe-off. In preparation for the over-extended hip angle, the subject forcibly maintained a dorsi-flexed ankle beyond peak stiffness (which occurs at 50\% of the gait cycle). When toe-off eventually occurred, the stiffness was thus insufficient to quickly restore the ankle to the plantar-flexed equilibrium.

\subsection{Comparison of dynamics}
\textit{Set A} and \textit{B} resulted in higher torque during terminal-stance in comparison to the other sets. This is likely due to higher dorsi-flexion in mid and terminal-stance (as discussed in Section \ref{kinematic}). As a consequence, the corresponding power was higher in \textit{Set A} and \textit{B}. Most interesting to note was the abrupt change in the torque corresponding to \textit{Set A} at foot-drop (13\%). Such a change was not observed in the results of the other sets. This is undoubtedly a consequence of the change in \textit{Set A}'s equilibrium angle at foot-drop. A similar behavior was observed at heel-off (40\%) in the results of both \textit{Set A} and \textit{B}. Thus, fewer changes in equilibrium angles ensure a smoother torque output. Further, the re-positioning of the ankle joint to the swing dorsi-flexed angle resulted in positive torque at the beginning of the swing phase for \textit{Sets A} to \textit{C}. In regards to the power curves, \textit{Set A}'s power output displays an aberrant increase at heel-off. The backing rationale is the high velocity arising from the hastened push-off detailed in Section \ref{kinematic}. Finally, the push-off power associated with \textit{Set D} was significantly lower due to the previously discussed delayed push-off.

\subsection{Limitations of this study}
The aforementioned height difference between the subject's limbs forced him to adopt a limp; i.e. a shorter step length and longer stance phase on the limb without the prosthetic. A major drawback of these gait asymmetries was insufficient loading of the prosthetic ankle during mid-stance; resulting in the reduced dorsi-flexion. Further, the swing equilibrium angle could have been tuned in a more systematic manner; i.e. the enforced swing dorsi-flexed angle could have been uniform across all sets. Finally, the usage of time-based control enforced the stringent requirement of gait synchronization on the subject.  

\section{\uppercase{Conclusion}}
This study proposed a least squares approach to estimating ankle impedance based on the work of \cite{Sup2008}. The resulting impedance values followed a trend consistent with perturbation studies \cite{Rouse2014,Lee2016a,Shorter2018}. The estimated impedance parameters were not sensitive to the number of equilibria enforced. On studying the effect of multiple equilibria on the impedance controller's performance, the following results were revealed: (i) Multiple equilibria during mid and terminal-stance phase that increase the plantar-flexed equilibrium angle (in increments) can assure more ankle dorsi-flexion during mid-stance. Consequently, the generated torque and power at push-off are higher (ii) Abrupt changes in torque can be expected at the instances when the equilibrium angle switches. Such changes impact the robustness of the system to perturbations. This ideology is the motivation behind studies in continuum of equilibria \cite{Mohammadi2019} (iii) While overly plantar-flexed equilibrium angle during terminal-stance results in higher push-off torque, it can give rise to premature push-off (iv) Most importantly, a single equilibrium angle during stance phase is sufficient to generate human-like kinematics and dynamics. 

\section{\uppercase{Future work}}
To overcome the prior listed limitations, a state-based control scheme will be implemented to enable flexibility in gait speed. The basis of such a control scheme can be found in \cite{Hong2019}. Current efforts are targeted at overcoming the limitations of the control scheme by using sensor-fusion. This paper assumed the order of the stiffness and damping polynomials to be the same during the stance phase. Future attempts will investigate the validity of this assumption. Presently, a height adjustable prosthesis is under development. This prosthesis will be used in all future studies of the impedance controller. Based on this study, an automated tuning algorithm will be developed and consecutively implemented on a transfemoral prosthesis. The dorsi-flexion observed during mid-stance will serve as the primary indication of well-tuned stiffness and damping parameters. Further, the impedance parameters proposed in this study will be used in a hybrid control strategy that would implement impedance control during stance phase followed by trajectory tracking during swing \cite{Hong2019}. 

\begin{figure}[t]
      \centering
      \framebox{\parbox{3.3in}{\centering\includegraphics[width = 3.25in]{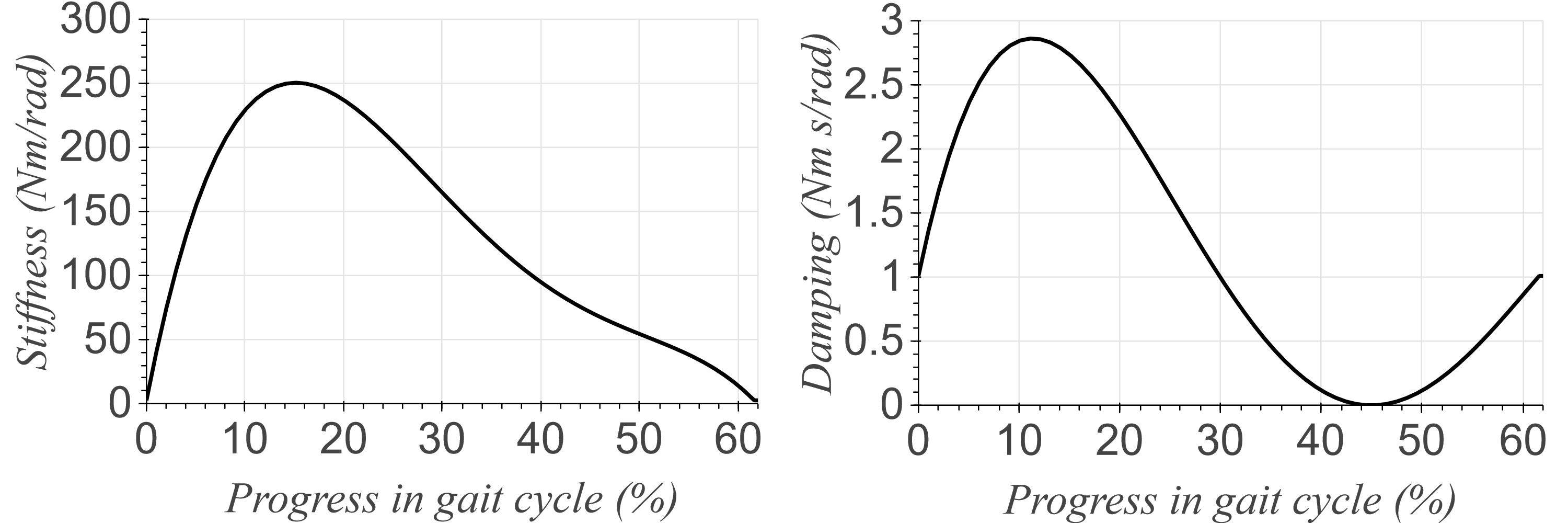}
        }}
      \caption{Preliminary estimation of knee impedance using the proposed theoretical approach. The estimation was limited to the stance phase of the gait cycle. \textbf{Left:} Stiffness, \textbf{Right:} Damping}
      \label{fig:KneeImp}
\end{figure}

In regards to the knee, a preliminary estimation of knee impedance during stance has been presented in Fig. \ref{fig:KneeImp}. The estimation utilized the sectioning proposed in \textit{Set B}. The optimized equilibrium angles were 0.1413 rad of knee flexion during initial stance, followed by 0.1968 rad until toe-off. The result is consistent with the values implemented by \cite{Sup2008} and \cite{Mohammadi2019}. 
Considering the extensive knee movement during swing, it may be more desirable to use trajectory tracking during swing. Another path worthy of investigation is the concept of a continuum of equilibria proposed by \cite{Mohammadi2019}.


\section*{APPENDIX}
The following figure depicts the fit attained using the least squares optimization. Also included is human torque data from \cite{Winter2009}. Table \ref{tab:polyCoeff} presents the optimized coefficients of the stiffness and damping polynomials.

\begin{figure}[h]
      \centering
      \framebox{\centering\includegraphics[width = 3.3in]{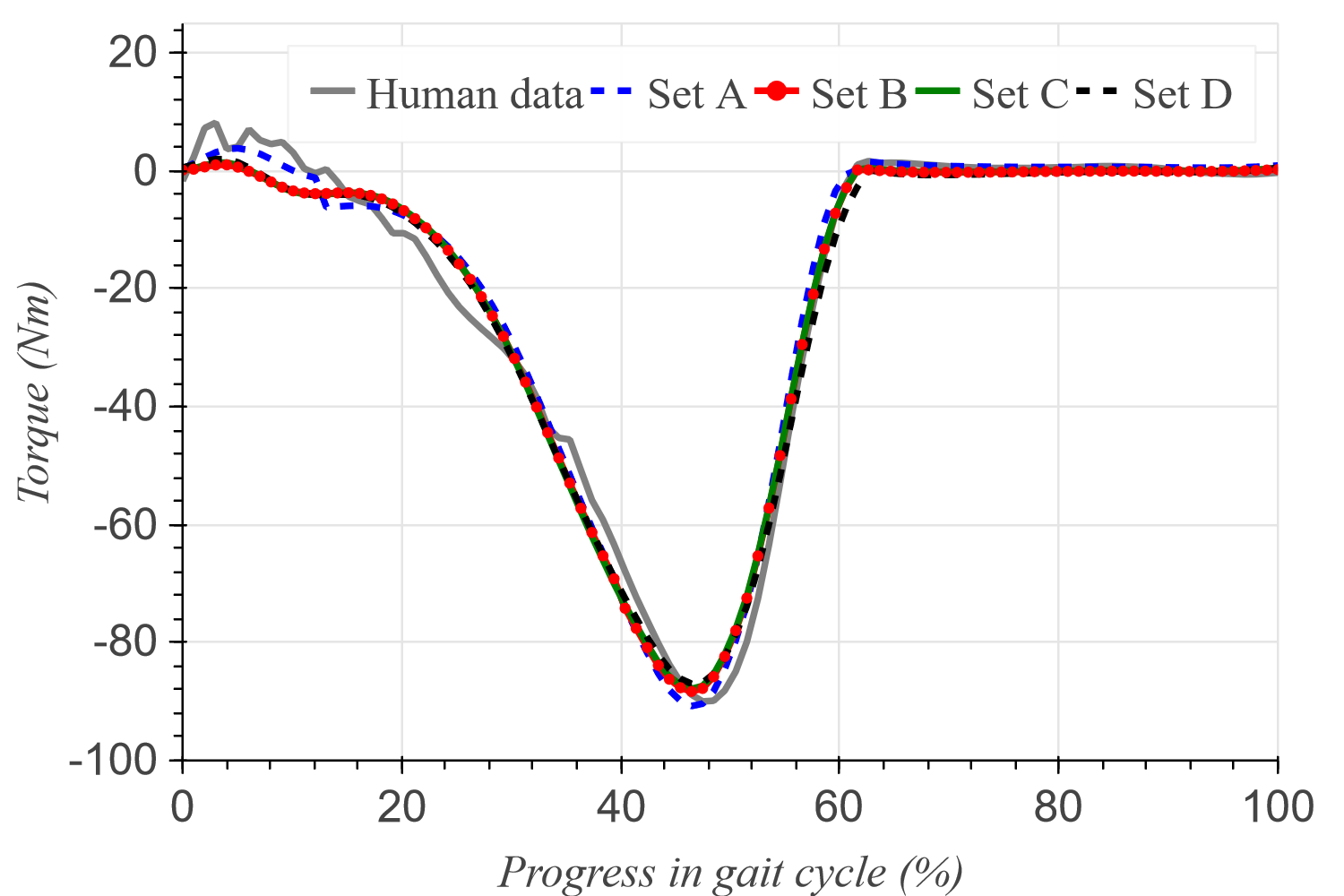}}
      \caption{The torque $\tau$ generated from the optimization in comparison to human torque data \cite{Winter2009}}
      \label{fig:torquefit}
\end{figure}

\begin{table}[h]
\caption{The coefficients of the polynomial curves}
\label{tab:polyCoeff}
\begin{center}
\begin{tabular}{c c c c c c}
\hline
\hline
\multicolumn{6}{c}{\textbf{Stiffness}} \\
\textbf{Set label}   & $k_4$  & $k_3$ & $k_2$   & $k_1$ & $k_0$\\
\hline
\hline
Set A   & -29870.57 & 28322.46 & -7061.82 & 586.04 & 2.21\\ 
Set B   & -19977.92 & 17340.71 & -3424.51 & 199.97 & 0.32\\ 
Set C   & -19822.71 & 17146.19 & -3333.05 & 181.16 & 0.75\\ 
Set D   & -16520.32 & 14144.17 & -2596.67 & 136.56 & 0.00\\
\hline
\hline
\multicolumn{6}{c}{\textbf{Damping}} \\
\textbf{Set label}   & $d_4$  & $d_3$ & $d_2$ & $d_1$  & $d_0$\\
\hline
\hline
Set A   & -22.45 & 88.08 & -76.20 & 18.76 & 0.12\\ 
Set B   & -140.21 & 261.35 & -158.46 & 31.21 & 0.12\\ 
Set C   & -164.32 & 303.05 & -181.22 & 35.04 & 0.18\\ 
Set D   & -171.23 & 311.36 & -182.97 & 34.53 & 0.26\\
\hline
\hline
\end{tabular}
\end{center}
\end{table}

\section*{ACKNOWLEDGMENT}

The authors thank Kenny Chour for his assistance during experimentation.

\addtolength{\textheight}{-13.0cm}   

\bibliographystyle{IEEEtran} 
\bibliography{root}

\begin{thebibliography}{10}
\providecommand{\url}[1]{#1}
\csname url@samestyle\endcsname
\providecommand{\newblock}{\relax}
\providecommand{\bibinfo}[2]{#2}
\providecommand{\BIBentrySTDinterwordspacing}{\spaceskip=0pt\relax}
\providecommand{\BIBentryALTinterwordstretchfactor}{4}
\providecommand{\BIBentryALTinterwordspacing}{\spaceskip=\fontdimen2\font plus
\BIBentryALTinterwordstretchfactor\fontdimen3\font minus
  \fontdimen4\font\relax}
\providecommand{\BIBforeignlanguage}[2]{{%
\expandafter\ifx\csname l@#1\endcsname\relax
\typeout{** WARNING: IEEEtran.bst: No hyphenation pattern has been}%
\typeout{** loaded for the language `#1'. Using the pattern for}%
\typeout{** the default language instead.}%
\else
\language=\csname l@#1\endcsname
\fi
#2}}
\providecommand{\BIBdecl}{\relax}
\BIBdecl

\bibitem{Perry1997}
J.~Perry, L.~A. Boyd, S.~S. Rao, and S.~J. Mulroy, ``Prosthetic weight
  acceptance mechanics in transtibial amputees wearing the single axis, seattle
  lite, and flex foot,'' \emph{IEEE Transactions on Rehabilitation
  Engineering}, vol.~5, no.~4, pp. 283--289, 1997.

\bibitem{Blumentritt1997}
S.~Blumentritt, H.~W. Scherer, U.~Wellershaus, and J.~W. Michael, ``Design
  principles, biomechanical data and clinical experience with a polycentric
  knee offering controlled stance phase knee flexion: a preliminary report,''
  \emph{JPO: Journal of Prosthetics and Orthotics}, vol.~9, no.~1, pp. 18--24,
  1997.

\bibitem{Romo2000}
D.~Romo, ``Prosthetic knees,'' \emph{Physical medicine and rehabilitation
  clinics of North America}, vol.~11, pp. 595--607, vii, 2000.

\bibitem{Au2007}
S.~K. {Au}, H.~{Herr}, J.~{Weber}, and E.~C. {Martinez-Villalpando}, ``Powered
  ankle-foot prosthesis for the improvement of amputee ambulation,'' in
  \emph{2007 29th Annual International Conference of the IEEE Engineering in
  Medicine and Biology Society}, Aug 2007, pp. 3020--3026.

\bibitem{Lenzi2017}
T.~{Lenzi}, M.~{Cempini}, J.~{Newkirk}, L.~J. {Hargrove}, and T.~A. {Kuiken},
  ``A lightweight robotic ankle prosthesis with non-backdrivable cam-based
  transmission,'' in \emph{2017 International Conference on Rehabilitation
  Robotics (ICORR)}, July 2017, pp. 1142--1147.

\bibitem{Martinez-Villalpando2008}
E.~C. {Martinez- Villalpando}, J.~{Weber}, G.~{Elliott}, and H.~{Herr},
  ``Design of an agonist-antagonist active knee prosthesis,'' in \emph{2008 2nd
  IEEE RAS EMBS International Conference on Biomedical Robotics and
  Biomechatronics}, Oct 2008, pp. 529--534.

\bibitem{Sup2008}
F.~Sup, A.~Bohara, and M.~Goldfarb, ``Design and control of a powered
  transfemoral prosthesis,'' \emph{The International Journal of Robotics
  Research}, vol.~27, no.~2, pp. 263--273, 2008.

\bibitem{Thatte2014}
N.~{Thatte} and H.~{Geyer}, ``Towards local reflexive control of a powered
  transfemoral prosthesis for robust amputee push and trip recovery,'' in
  \emph{2014 IEEE/RSJ International Conference on Intelligent Robots and
  Systems}, Sep. 2014, pp. 2069--2074.

\bibitem{Windrich2016}
M.~Windrich, M.~Grimmer, O.~Christ, S.~Rinderknecht, and P.~Beckerle, ``Active
  lower limb prosthetics: a systematic review of design issues and solutions,''
  \emph{Biomedical Engineering Online}, vol.~15, no.~3, p. 140, 2016.

\bibitem{Zhao2017a}
H.~Zhao, J.~Horn, J.~Reher, V.~Paredes, and A.~D. Ames, ``{First steps toward
  translating robotic walking to prostheses: a nonlinear optimization based
  control approach},'' \emph{Autonomous Robots}, vol.~41, no.~3, pp. 725--742,
  2017.

\bibitem{Azimi2017}
V.~{Azimi}, T.~{Shu}, H.~{Zhao}, E.~{Ambrose}, A.~D. {Ames}, and D.~{Simon},
  ``Robust control of a powered transfemoral prosthesis device with
  experimental verification,'' in \emph{2017 American Control Conference
  (ACC)}, May 2017, pp. 517--522.

\bibitem{Elery2018}
T.~{Elery}, S.~{Rezazadeh}, C.~{Nesler}, J.~{Doan}, H.~{Zhu}, and R.~D.
  {Gregg}, ``Design and benchtop validation of a powered knee-ankle prosthesis
  with high-torque, low-impedance actuators,'' in \emph{2018 IEEE International
  Conference on Robotics and Automation (ICRA)}, May 2018, pp. 2788--2795.

\bibitem{Azocar2018}
A.~F. {Azocar}, L.~M. {Mooney}, L.~J. {Hargrove}, and E.~J. {Rouse}, ``Design
  and characterization of an open-source robotic leg prosthesis,'' in
  \emph{2018 7th IEEE International Conference on Biomedical Robotics and
  Biomechatronics (Biorob)}, Aug 2018, pp. 111--118.

\bibitem{Fey2014}
N.~P. Fey, A.~M. Simon, A.~J. Young, and L.~J. Hargrove, ``Controlling knee
  swing initiation and ankle plantarflexion with an active prosthesis on level
  and inclined surfaces at variable walking speeds,'' \emph{IEEE Journal of
  Translational Engineering in Health and Medicine}, vol.~2, pp. 1--12, 2014.

\bibitem{Gregg2014}
R.~D. Gregg, T.~Lenzi, L.~J. Hargrove, and J.~W. Sensinger, ``{Virtual
  constraint control of a powered prosthetic leg: From simulation to
  experiments with transfemoral amputees},'' \emph{IEEE Transactions on
  Robotics}, vol.~30, no.~6, pp. 1455--1471, 2014.

\bibitem{Paredes2016}
V.~{Paredes}, W.~{Hong}, S.~{Patrick}, and P.~{Hur}, ``Upslope walking with
  transfemoral prosthesis using optimization based spline generation,'' in
  \emph{2016 IEEE/RSJ International Conference on Intelligent Robots and
  Systems (IROS)}, Oct 2016, pp. 3204--3211.

\bibitem{Martin2017}
A.~E. Martin and R.~D. Gregg, ``Stable, robust hybrid zero dynamics control of
  powered lower-limb prostheses,'' \emph{IEEE Transactions on Automatic
  Control}, vol.~62, no.~8, pp. 3930--3942, 2017.

\bibitem{Winter2009}
D.~A. Winter, \emph{{Biomechanics and motor control of human movement}},
  4th~ed.\hskip 1em plus 0.5em minus 0.4em\relax Wiley, 2009.

\bibitem{Eilenberg2010}
M.~F. Eilenberg, H.~Geyer, and H.~Herr, ``{Control of a Powered Ankle–Foot
  Prosthesis Based on a Neuromuscular Model},'' \emph{IEEE Transactions on
  Neural Systems and Rehabilitation Engineering}, vol.~18, no.~2, pp. 164--173,
  2010.

\bibitem{Blaya2004}
J.~Blaya and H.~Herr, ``{Adaptive Control of a Variable-Impedance Ankle-Foot
  Orthosis to Assist Drop-Foot Gait},'' \emph{IEEE Transactions on Neural
  Systems and Rehabilitation Engineering}, vol.~12, no.~1, pp. 24--31, 2004.

\bibitem{Shamaei2013b}
K.~Shamaei, G.~S. Sawicki, and A.~M. Dollar, ``{Estimation of Quasi-Stiffness
  of the Human Knee in the Stance Phase of Walking},'' \emph{PLOS ONE}, vol.~8,
  no.~3, 2013.

\bibitem{Hu2016}
Y.~Hu and K.~Mombaur, ``{Analysis of human leg joints compliance in different
  walking scenarios with an optimal control approach},''
  \emph{IFAC-PapersOnLine}, vol.~49, no.~14, pp. 99--106, 2016.

\bibitem{Rouse2014}
E.~J. Rouse, L.~J. Hargrove, E.~J. Perreault, and T.~A. Kuiken, ``{Estimation
  of Human Ankle Impedance During the Stance Phase of Walking},'' \emph{IEEE
  Transactions on Neural Systems and Rehabilitation Engineering}, vol.~22,
  no.~4, pp. 870--878, 2014.

\bibitem{Lee2016a}
H.~Lee, E.~J. Rouse, and H.~I. Krebs, ``{Summary of Human Ankle Mechanical
  Impedance During Walking},'' \emph{IEEE Journal of Translational Engineering
  in Health and Medicine}, vol.~4, pp. 1--7, 2016.

\bibitem{Shorter2018}
A.~L. Shorter and E.~J. Rouse, ``{Mechanical Impedance of the Ankle during the
  Terminal Stance Phase of Walking},'' \emph{IEEE Transactions on Neural
  Systems and Rehabilitation Engineering}, vol.~26, no.~1, pp. 135--143, 2018.

\bibitem{Koopman2016}
B.~Koopman, E.~H. {Van Asseldonk}, and H.~{Van Der Kooij}, ``{Estimation of
  Human Hip and Knee Multi-Joint Dynamics Using the LOPES Gait Trainer},''
  \emph{IEEE Transactions on Robotics}, vol.~32, no.~4, pp. 920--932, 2016.

\bibitem{Mohammadi2019}
A.~Mohammadi and R.~D. Gregg, ``Variable impedance control of powered knee
  prostheses using human-inspired algebraic curves,'' \emph{Journal of
  Computational and Nonlinear Dynamics}, vol.~14, pp. 1--10, 2019.

\bibitem{Hong2019}
W.~{Hong}, V.~{Paredes}, K.~{Chao}, S.~{Patrick}, and P.~{Hur}, ``Consolidated
  control framework to control a powered transfemoral prosthesis over inclined
  terrain conditions,'' in \emph{2019 International Conference on Robotics and
  Automation (ICRA)}, May 2019, pp. 2838--2844.

\bibitem{Rezazadeh2018}
S.~{Rezazadeh}, D.~{Quintero}, N.~{Divekar}, and R.~D. {Gregg}, ``A phase
  variable approach to volitional control of powered knee-ankle prostheses,''
  in \emph{2018 IEEE/RSJ International Conference on Intelligent Robots and
  Systems (IROS)}, Oct 2018, pp. 2292--2298.

\bibitem{Rosenblatt2014}
N.~J. Rosenblatt, A.~Bauer, D.~Rotter, and M.~D. Grabiner, ``Active
  dorsiflexing prostheses may reduce trip-related fall risk in people with
  transtibial amputation,'' \emph{Journal of Rehabilitation Research \&
  Development}, vol.~51, no.~8, 2014.

\end{thebibliography}


\end{document}